# Artificial Immune Systems (AIS) – A New Paradigm for Heuristic Decision Making

Uwe Aickelin

School of Computer Science and Information Technology
The University of Nottingham
Nottingham, NG8 1BB, United Kingdom
`uxa@cs.nott.ac.uk`

**Abstract**
Over the last few years, more and more heuristic decision making techniques have been inspired by nature, e.g. evolutionary algorithms, ant colony optimisation and simulated annealing. More recently, a novel computational intelligence technique inspired by immunology has emerged, called Artificial Immune Systems (AIS). This immune system inspired technique has already been useful in solving some computational problems. In this keynote, we will very briefly describe the immune system metaphors that are relevant to AIS. We will then give some illustrative real-world problems suitable for AIS use and show a step-by-step algorithm walkthrough. A comparison of AIS to other well-known algorithms and areas for future work will round this keynote off. It should be noted that as AIS is still a young and evolving field, there is not yet a fixed algorithm template and hence actual implementations might differ somewhat from the examples given here.

## Brief Overview of the Natural Immune System

The biological immune system is an elaborate defence mechanism which has evolved over millions of years through extensive redesigning, testing, tuning and optimisation processes. While many details of the immune mechanisms and processes are yet unknown even to immunologists, it is, however, well-known that the immune system uses a multi-level defence both in parallel and sequential fashion. Depending on the type of the intruder (also called antigen or pathogen), and the way it gets into the body, the immune system uses different response mechanisms either to neutralize the pathogenic effect or to destroy the infected cells. A detailed overview of the immune system can be found in many textbooks[13,16]. The immune features that are particularly relevant to this keynote are matching, diversity and distributed control. Matching refers to the binding between antibodies and antigens. Diversity refers to the fact that, in order to achieve optimal antigen space coverage, antibody diversity must be encouraged[10]. Distributed control means that there is no central controller. Rather, the immune system is governed by local interactions among immune cells and antigens.

As mentioned above, the human body is protected against foreign invaders by a multi-layered system. The immune system is composed of physical barriers such as the skin and respiratory system, physiological barriers such as destructive enzymes and stomach acids and the actual immune system, which has two complementary parts, the innate and adaptive immune systems. The innate immune system is an unchanging mechanism that detects and destroys certain invading organisms, whilst the adaptive (or acquired) immune system responds to previously unknown foreign cells and builds a response that can remain in the body over a long period of time.

Of most interest to us is the adaptive immune system, which is composed of a number of different agents performing different functions at a number of different locations in the body. The precise interaction of these agents is still a topic for debate[13,16]. Two of the most important cells in this process are two types of white blood cells, called T-cells and B-cells. Both of these originate in the bone marrow (hence the 'B'), but T-cells pass on to the thymus to mature (hence the 'T'), before they circulate the body in the blood and lymphatic vessels.

T-cells come in three types; T-helper cells which are essential to the activation of B-cells, Killer T-cells which bind to foreign invaders and inject poisonous chemicals into them causing their destruction, and suppressor T-cells which inhibit the action of other immune cells thus preventing allergic reactions and autoimmune diseases.

B-cells are responsible for the production and secretion of antibodies, which are specific proteins that bind to the antigen. Each B-cell can only produce one particular antibody. The antigen is found on the surface of the invading organism and the binding of an antibody to the antigen is a signal to destroy the invading cell.

As can be gleaned from the brief explanations above, there is more than one mechanism at work in the human immune system[8,12,13,16]. However let us now concentrate on the essential process exploited in most AIS: The matching between antigen and antibody which subsequently leads to increased concentrations (proliferation) of more closely matched antibodies. In particular, the negative selection mechanism and the 'clonal selection' and 'somatic hypermutation' theories are primarily used in AIS models.

*Negative Selection mechanism*
The purpose of negative selection is to provide tolerance for self cells. It deals with the immune system's ability to detect unknown antigens whilst not reacting to self. During the generation of T-cells, receptors are made through a pseudo-random genetic rearrangement process. Then, they undergo a censoring process in the thymus (hence the 'T' in T-Cell), called negative selection. There, T-cells that react against self-proteins are destroyed and thus, only those that do not bind to self-proteins are allowed to leave the thymus. These matured T-cells then circulate throughout the body to perform immunological functions and protect the body against foreign antigens.

*Clonal Selection Principle*
The clonal selection principle describes the basic features of an immune response to an antigenic stimulus. It establishes the idea that only those cells that recognize the antigen proliferate, thus being selected against those that do not. The main features of the clonal selection theory are that:
- New cells are (cloned) copies of their parents, subject to a mutation mechanism (somatic hypermutation);
- Self-reactive cells are eliminated;
- Proliferation and differentiation of mature cells on contact with antigens.

When an antibody strongly matches an antigen the corresponding B-cell is stimulated to produce clones of itself that then produce more antibodies. This (hyper)mutation is quite rapid, often as much as "one mutation per cell division"[7] to allow a very quick response to antigens. It should be noted here that the AIS literature often makes no distinction between B-cells and the antibodies they produce. Both are subsumed under the word 'antibody' and statements such as mutation of antibodies (rather than mutation of B-cells) are common. For simplicity, we will only use the term 'antibody' in the following sections.

There are many more features of the immune system, including the immune network theory[12], adaptation, immunological memory and protection against auto-immune attacks, not discussed here. In the following sections, we will revisit some important aspects of these concepts and show how they can be modelled in 'artificial' immune systems used to solve real-world problems. First, let us give an overview of typical problems that we believe are amenable to being solved by AIS.

## Illustrative problems
*Data Mining – Collaborative Filtering (CF) and Clustering*
CF is a term for a broad range of algorithms that use similarity measures to obtain information, usually recommendations. The best-known example is probably the "people who bought this also bought that" feature of the internet company Amazon[3]. However, any problem domain where users are required to rate items is amenable to CF techniques. Commercial applications are usually called recommender systems[15,17]. A canonical example is film recommendation[4,5], which we will use in this keynote.

In traditional CF, the items to be recommended are treated as 'black boxes'. That is, your recommendations are based purely on the votes of other users, and not on the content of the item. The preferences of a user, usually a set of votes on an item, comprise a user profile, and these profiles are compared in order to build a neighbourhood. The key decision is what similarity measure is used. The most common method to compare two users are correlation-based measures like Pearson or Spearman, which give two neighbours a matching score between -1 and 1. The canonical example is the k-Nearest-Neighbour algorithm, which uses a matching method to select *k* reviewers with high similarity measures. The votes from these reviewers, suitably weighted, are used to make predictions and recommendations.

The evaluation of a CF algorithm usually centres on its accuracy. There is a difference between prediction (given a film, predict a given user's rating of that film) and recommendation (given a user, suggest films that are likely to attract a high rating). Prediction is easier to assess quantitatively but recommendation is a more natural fit to the film domain. A related problem to CF is that of clustering data or users in a database. This is particularly useful in huge databases, which have become too large to handle. Clustering works by dividing the entries of the database into groups, which contain people with similar preferences or in general data of similar type.

*Computer Security - Intrusion Detection Systems (IDS)*
Anyone keeping up-to-date with current affairs in computing can confirm numerous cases of attacks made on computer servers of well-known companies. These attacks range from denial-of-service to extracting credit-card details and one wonders "haven't they installed any security such as a firewall"? The fact is they often have a firewall. A firewall is useful, often essential, but current firewall technology is insufficient to detect and block all kinds of attacks.

However, on ports that need to be open to the internet, a firewall can do little to prevent attacks. Moreover, even if a port is blocked from internet access, this does not stop an attack from inside the organisation. This is where IDS come in. As the name suggests, IDS are installed to identify (potential) attacks and to react by usually generating an alert or blocking the unscrupulous data.

The main goal of IDS is to detect unauthorised use, misuse and abuse of computer systems by both system insiders and external intruders. Most current IDS define suspicious signatures based on known intrusions and probes. The obvious limit of this type of IDS is its failure of detecting previously unknown intrusions. In contrast, the human immune system adaptively generates new immune cells so that it is able to detect previously unknown and rapidly evolving harmful antigens[9]. Thus, the challenge is to emulate the success of the natural systems to protect computers.

## Basic AIS Concepts
*Initialisation / Encoding*
Along with other heuristics, choosing a suitable encoding is very important for the algorithm's success. Similar to Genetic Algorithms, there is close inter-play between the encoding and the fitness function (in AIS referred to as the 'matching' or 'affinity' function). Hence, both ought to be thought about at the same time. For the current discussion, let us begin with the encoding.

First, let us define what is meant by 'antigen' and 'antibody' in the context of an application domain. Typically, an antigen is the 'target', e.g. the data item to be checked to see if it is an intrusion, or the user to be clustered or made a recommendation for. The antibodies are the remainder of the data, e.g. other users in the database, general network traffic that has already been identified etc. Sometimes, there can be more than one antigen at a time and there are usually a large number of antibodies present simultaneously.

Antigens and antibodies are represented or encoded in the same way. For most problems the most obvious representation is a string of numbers or features, where the length is equal to the number of variables, the position is the variable identifier and the value is the actual value of the variable itself (e.g. binary or real). For instance, in a five variable binary problem, an encoding could look like this: (10010).

Previously we mentioned data mining and intrusion detection applications. What would an encoding look like in these cases? For data mining, let us consider the problem of recommending films. Here the encoding has to represent a user's profile with regards to the films he has seen and how much he has (dis)liked them. A possible encoding for this could be a list of numbers, where each number represents the 'vote' for an item. Votes could be binary (e.g. Did you see the film?), but can also be integers in a range (say [0, 5], i.e. 0 - did not like the film at all, 5 – did like the film very much).

Hence for film recommendation, a possible encoding is:

$$User = \{\{id_1, score_1\}, \{id_2, score_2\}...\{id_n, score_n\}\}$$

Where *id* corresponds to the unique identifier of the film being rated and score to this user's score for that film. This captures the essential features of the data available[4,5].

For intrusion detection, the encoding needs to encapsulate the essence of each data packet transferred, e.g. [<protocol> <source ip> <source port> <destination ip> <destination port>], for example: [<tcp> <113.112.255.254> <108.200.111.12> <25>] which represents an incoming data packet sent to port 25. In these scenarios, wildcards like 'any port' are also often used.

*Similarity or Affinity Measure (Fitness Function)*
As mentioned above, similarity measures or matching rules are very important design choices in developing an AIS algorithm, and closely coupled to the encoding scheme.

Two of the simplest matching algorithms are best explained using binary encoding: Consider the strings (00000) and (00011). If one does a bit-by-bit comparison, the first three bits are identical and hence one could give this pair a matching score of 3. In other words, one computes the opposite of the Hamming Distance (which is defined as the number of bits that have to be changed in order to make the two strings identical).

Now consider this pair: (00000) and (01010). Again, simple bit matching gives us a similarity score of 3. However, the matching is quite different as the three matching bits are not connected. Depending on the problem and encoding, this might be better or worse. Thus, another simple matching algorithm is to count the number of continuous bits that match and return the length of the longest matching as the similarity measure.

For the first example above, this would still be 3, for the second example this would be 1.

If the encoding is non-binary, e.g. real variables, there are even more possibilities to compute the 'distance' between the two strings, for instance one could compute the geometrical (Euclidian) distance etc.

For data mining problems, similarity often means 'correlation'. Take the film recommendation problem as an example and assume that we are trying to find users in a database that are similar to the target user whose profile we are trying to match to make recommendations. In this case, what we are trying to measure is how similar the two users' tastes are. One of the easiest ways of doing this is to compute the Pearson Correlation Coefficient between the two users, e.g. the Pearson measure is used to compare two users $u$ and $v$:

$$r = \frac{\sum_{i=1}^{n}(u_i - \bar{u})(v_i - \bar{v})}{\sqrt{\sum_{i=1}^{n}(u_i - \bar{u})^2 \sum_{i=1}^{n}(v_i - \bar{v})^2}}$$

Where $u$ and $v$ are users, $n$ is the number of overlapping votes (i.e. films for which both $u$ and $v$ have voted), $u_i$ is the vote of user $u$ for film $i$ and $\bar{u}$ is the average vote of user u over all films (not just the overlapping votes). During previous research[4,5], it was found useful to introduce a penalty parameter (c.f. penalties in genetic algorithms) for users who only have very few films in common, reducing their correlation which might be artificially high.

The outcome of this measure is a value between -1 and 1, where values close to 1 mean strong agreement, values near to -1 mean strong disagreement and values around 0 mean no correlation. From a data mining point of view, those users who score either 1 or -1 are the most useful and hence will be selected for further treatment by the algorithm. This approach is called 'positive selection' and might be coupled to some form of mutation as explained in the 'clonal selection' section above. For some problem domains, such as the film recommender, mutation might not make any sense and hence pure cloning is used.

In other applications domains, 'matching' might not actually be beneficial and hence we might want to eliminate those items that match. This approach is known as 'negative selection'. Under what circumstance would a negative selection algorithm be suitable for an AIS implementation? Consider IDS as used by Hofmeyr and Forrest[11]. One way of solving this problem is by defining a set of 'self', i.e. a trusted network, our company's computers, known partners etc. During the initialisation of the algorithm, we then randomly create a large number of so called 'detectors' strings. We then subject these detectors to a matching algorithm that compares them to our 'self'. Any matching detectors are eliminated and hence those that do no match are selected and kept (negative selection). Eventually, these detectors form a final set which is then used in the second phase of the algorithm to continuously monitor all network traffic. Should a match be found now, the algorithm reports this as a possible alert of 'non-self'.

*Negative or Clonal Selection*
The meaning of 'selection' differs somewhat depending on the exact problem the AIS is applied to. We have already briefly described the concept of negative and clonal selection earlier. For the film recommender, choosing a suitable neighbourhood of users means choosing good correlation scores and hence we will perform 'positive' selection. How would the algorithm do this?

Consider the AIS to be empty at the beginning. The target user is encoded as the antigen, and all other users in the database are possible antibodies. We add the antigen to the AIS and then we add one candidate antibody at a time. Antibodies will start with a certain concentration value. This value represents the natural lifespan of antibodies and decreases over time (death rate), similar to the evaporation in Ant Systems. Antibodies with a sufficiently low concentration are removed from the system, whereas antibodies with a high concentration may saturate. An antibody can increase its concentration by matching the antigen: The better the match the higher the increase (a process called 'stimulation'). The process of stimulation or increasing concentration can also be regarded as 'cloning' if one thinks in a discrete setting. Once enough antibodies have been added to the system, it starts to iterate a loop of suppression and stimulation until at least one antibody drops out. A new antibody is then added and the process is repeated until the AIS has stabilised, i.e. until there are no more drop-outs for a certain period of time.

Mathematically, in each step (iteration) an antibody's concentration is increased by an amount dependent on its matching to the antigen. In the absence of matching, an antibody's concentration will slowly decrease over time. Hence, an AIS iteration is governed by the following equation[8]:

$$\frac{dx_i}{dt} = \left[ \begin{pmatrix} antigens \\ recognised \end{pmatrix} - \begin{pmatrix} death \\ rate \end{pmatrix} \right] = \left[ k_2 (\sum_{j=1}^{N} m_{ji} x_i y_j) - k_3 x_i \right]$$

Where:
N is the number of antigens.
$x_i$ is the concentration of antibody we
$y_j$ is the concentration of antigen j
$k_2$ is the stimulation effect and $k_3$ is the death rate
$m_{ji}$ is the matching function between antibody we and antibody (or antigen) j

The following pseudo code summarise the AIS film recommender:

```
Initialise AIS
Encode user for whom to make predictions as antigen AG
WHILE (AIS not Full) & (More Antibodies) DO
        Add next user as an antibody AB
        Calculate matching scores between AB and AG
        WHILE (AIS at full size) & (AIS not Stabilised) DO
                Reduce Concentration of all ABs by a fixed amount
                Match each AB against AG and stimulate as necessary
        OD
OD
Use final set of Antibodies to produce recommendation.
```

In this example, the AIS is considered stable after iterating ten times without changing in size. Stabilisation thus means that a sufficient number of 'good' neighbours have been identified and therefore a prediction can be made. 'Poor' neighbours would be expected to drop out of the AIS after a few iterations. Once the AIS has stabilised using the above algorithm, we use the antibody concentration to weigh the neighbours and then perform a weighted average type recommendation.

*Somatic Hypermutation (optional)*
The mutation most commonly used in AIS is very similar to that found in Genetic Algorithms, e.g. for binary strings bits are flipped, for real value strings one value is changed at random, or for others the order of elements is swapped. In addition, the mechanism is often enhanced by the 'somatic' idea, i.e. the closer the match (or the less close the match, depending on what we are trying to achieve), the more (or less) disruptive the mutation.

However, mutating the data might not make sense for all problems considered. For instance, it would not be suitable for the film recommender. Certainly, mutation could be used to make users more similar to the target, however, the validity of recommendations based on these artificial users is questionable and if over-done, one could end up with the target user itself. Hence for some problems, Somatic Hypermutation is not used, since it is not immediately obvious how to mutate the data sensibly such that these artificial entities still represent plausible data.

Nevertheless, for other problem domains, mutation might be very useful. For instance, taking the negative selection approach to intrusion detection, rather than throwing away matching detectors in the first phase of the algorithm, these could be mutated to save time and effort. Also, depending on the degree of matching the mutation could be more or less strong. This was in fact one extension implemented by Hofmeyr and Forrest[11].

For data mining problems, mutation might also be useful, if for instance the aim is to cluster users. Then the centre (antibody) of each cluster could be an artificial pseudo user that can be mutated at will until the desired degree of matching between the centre and antigens in its cluster is reached[7].

## Comparison of AIS to Genetic Algorithms and Neural Networks

Going through the keynote so far, you might already have noticed that concepts similar to those in Genetic Algorithms and Neural Networks have been mentioned a number of times. In fact, both have a number of ideas in common with AIS and the purpose of the following, self-explanatory table, is to put their similarities and differences next to each other.

Evolutionary computation shares many concepts of AIS like a population, genotype phenotype mapping, and proliferation of the most fit. On the other hand, AIS models based on immune networks resemble the structures and interactions of connectionist (neural network) models. Some works have pointed out the similarities and the differences between AIS and other heuristics[6,7].

It should be noted that some of the items in Table 1 are gross simplifications, both to benefit the design of the table and not to overwhelm the reader. Many of these points are debatable; however, we believe that this comparison is valuable nevertheless to show approximately where AIS fit in. The comparisons are based on a Genetic Algorithm (GA) used for optimisation and a Neural Network (NN) used for classification.

|  | GA (Optimisation) | NN (Classification) | AIS |
|---|---|---|---|
| **Components** | Chromosome Strings | Artificial Neurons | Attribute Strings |
| **Location of Components** | Dynamic | Pre-Defined | Dynamic |
| **Structure** | Discrete Components | Networked Components | Discrete Components |
| **Knowledge Storage** | Chromosome Strings | Connection Strengths | Component Concentration |
| **Dynamics** | Evolution | Learning | Evolution & Learning |
| **Meta-Dynamics** | Recruitment & Elimination of Components | Construction & Pruning of Connections | Recruitment & Elimination of Components |
| **Interaction of Components** | Crossover | Network Connections | Recognition |
| **Interaction with Environment** | Fitness Function | External Stimuli | Recognition Function |

Table 1: Comparison of AIS to Genetic Algorithms and Neural Networks.

## Conclusions

The immune system is highly distributed, highly adaptive, self-organising, maintains a memory of past encounters and has the ability to continually learn about new encounters. AIS are systems developed around the current understanding of the immune system. We have tried to illustrate how an AIS can capture the basic elements of the immune system and exhibit some of its chief characteristics. More complex AIS can incorporate further properties of natural immune systems, including diversity, distributed computation, error tolerance, dynamic learning, adaptation and self-monitoring.

AIS are a general framework for a distributed adaptive system and could, in principle, be applied to many domains. AIS have been applied to classification problems, optimisation tasks and other domains. AIS share similarities with other biologically inspired systems and may be described as somewhat of a cross between genetic algorithms and neural networks.

An advantage over genetic algorithms is that in AIS (idiotypic) network effects[4,12] can easily be implemented. Advantages over neural networks are the benefits of a population of solutions and the evolutionary selection pressure and mutation. Other advantages of AIS are that only positive examples are required (e.g. 'self') and the patterns learnt can be explicitly examined. In addition, because AIS are self-organizing, not many system parameters are required.

To me, the attraction of the (artificial) immune system is that an adaptive pool of antibodies can produce 'intelligent' behaviour through a combination of evolution and idiotypic effects. In particular, a new and still controversial 'Danger Theory'[14] seems promising in this respect. This can be harnessed to tackle the problem of preference matching and current challenges in intrusion detection beyond the capabilities of current systems[2,1].

**References**
1. Aickelin U and Cayzer S (2002): The Danger Theory and Its Application to AIS, Proceedings 1st International Conference on AIS, pp 141-148, Canterbury, UK.
2. Aickelin U, Bentley P, Cayzer S, Kim J and McLeod J (2003): DT: The Link between AIS and IDS?, Proceedings 2nd International Conference on AIS, pp 147-155, Springer, Edinburgh, UK.
3. Amazon (2003), Amazon.com Recommendations, http://www.amazon.com/.
4. Cayzer S and Aickelin U (2002a), A Recommender System based on the Immune Network, in Proceedings CEC2002, pp 807-813, Honolulu, USA.
5. Cayzer S and Aickelin U (2002b), On the Effects of Idiotypic Interactions for Recommendation Communities in AIS, Proceedings 1st International Conference on AIS, pp 154-160, Canterbury, UK.
6. Dasgupta, D (1999), Artificial Immune Systems and Their Applications, Springer Verlag, 1999.

7. De Castro L and Von Zuben F (2001), Learning and Optimization Using the Clonal Selection Principle. IEEE Transactions on Evolutionary Computation, Special Issue on Artificial Immune Systems, 6(3), pp. 239-251.
8. Farmer J, Packard N and Perelson A (1986), The immune system, adaptation, and machine learning, Physica, vol. 22, pp. 187-204, 1986.
9. Forrest S, Perelson A, Allen L and Cherukuri R. Self-Nonself Discrimination in a Computer. In Proceedings of IEEE Symposium on Research in Security and Privacy, pp 202-212, Oakland, May 16-18 1994.
10. Hightower RR, Forrest S and Perelson AS (1995). The evolution of emergent organization in immune system gene libraries, Proceedings of the 6th Conference on Genetic Algorithms, pp. 344-350, 1995.
11. Hofmeyr S, Forrest S (2000), Architecture for an AIS, Evolutionary Computation, Vol. 7, No. 1, pp 1289-1296.
12. Jerne NK (1973), Towards a network theory of the immune system Annals of Immunology, vol. 125, no. C, pp. 373-389, 1973.
13. Kubi J (2002), Immunology, Fifth Edition by Richard A. Goldsby, Thomas J. Kindt, Barbara A. Osborne, W H Freeman.
14. Matzinger P (2001), The Danger Model in Its Historical Context, Scandinavian Journal of Immunology, 54: 4-9, 2001.
15. Morrison T and Aickelin U (2002): An AIS as a Recommender System for Web Sites, in Proceedings of the 1st International Conference on AIS (ICARIS-2002), pp 161-169, Canterbury, UK.
16. Perelson AS and Weisbuch G (1997), Immunology for physicists Reviews of Modern Physics, vol. 69, pp. 1219-1267, 1997.
17. Resnick P and Varian HR (1997), Recommender systems Communications of the ACM, vol. 40, pp. 56-58, 1997.